\pdfoutput=1

\documentclass[runningheads,a4paper]{llncs}

\usepackage[utf8]{inputenc}
\usepackage{llncsdoc}
\usepackage[english]{babel}
\usepackage{amsmath}
\usepackage{amssymb}
\usepackage{color}
\usepackage{algorithm2e}
\usepackage{graphicx}

\usepackage[super]{natbib}
\makeatletter
\renewcommand\bibsection%
{
	\section*{\refname
		\@mkboth{\MakeUppercase{\refname}}{\MakeUppercase{\refname}}}
}
\makeatother

\begin{document}

\title{A Theory of Diagnostic Interpretation in Supervised Classification}
\titlerunning{Diagnostic Interpretation}  
%
\author{Anirban Mukhopadhyay}

%
\authorrunning{Mukhopadhyay} 
\authorrunning{Anonymous} 

\institute{Department of Computer Science, TU Darmstadt\\
\email{anirban.mukhopadhyay@gris.tu-darmstadt.de}
}

\maketitle              

\begin{abstract}
Interpretable deep learning is a fundamental building block towards safer AI, especially when the deployment possibilities of deep learning-based computer-aided medical diagnostic systems are so eminent~\cite{Reuter}.
However, without a computational formulation of black-box interpretation, general interpretability research rely heavily on subjective bias.
Clear decision structure of the medical diagnostics lets us approximate the decision process of a radiologist as a model - removed from subjective bias.
We define the process of interpretation as a finite communication between a known model and a black-box model to optimally map the black box's decision process in the known model.
Consequently, we define interpretability as maximal information gain over the initial uncertainty about the black-box's decision within finite communication.  
We relax this definition based on the observation that diagnostic interpretation is typically achieved by a process of minimal querying.
We derive an algorithm to calculate diagnostic interpretability. 
The usual question of accuracy-interpretability tradeoff, i.e. whether a black-box model's prediction accuracy is dependent on its ability to be interpreted by a known source model, does not arise in this theory.   
With multiple example simulation experiments of various complexity levels, we demonstrate the working of such a theoretical model in synthetic supervised classification scenarios. 

\end{abstract}

\section{Introduction}\label{intro}
Reliable and accurate Computer-Aided medical Diagnostic (CAD) system is a long overdue - unprecedented accuracy achieved by Deep Learning techniques in disease diagnosis (compared to previous attempts of simplistic learning) is certainly inspiring in this matter.
Typical deep learning based CAD systems treat diagnosis as a supervised classification problem where expert annotated data is used for training the model and accuracy is measured by the deep learning model's performance on previously unseen examples.
In fact, strong diagnostic accuracy of Deep learning is recently demonstrated in multiple areas. 
However, in terms of classification, unpredictable errors made by Deep Learning under minor modification of input~\cite{szegedy2013intriguing} has already been identified.
In medicine, where trust-ability and traceability are gold standards for acceptance, whether some form of deep-learning will succeed as a go-to clinical CAD system in foreseeable future, depends mainly on solving~\emph{interpretability}.
To date, however, little work has thoroughly examined the \emph{Diagnostic Interpretability} of deep learning in CAD.

Interpretability is a relatively new and unexplored field of research especially in the deep learning context.
Lipton~\cite{lipton2016mythos} has written an article on the desiderata of interpretability. 
Recent trends of interpretability research has also been summarized by Velez and Kim~\cite{doshi2017towards} with a view about the prospects of making the general interpretability research a scientific discipline. 
We are mainly interested in the special case of \emph{diagnostic interpretability} - particularly for CAD in supervised learning setting.
Most of the dominant literature in diagnostic interpretability either presents visual interpretation of the deep network's decision process to radiologists, in form of some heat maps~\cite{garcia2018interpretable} or embeds textual description of the decision process as a proxy of deep network's decision~\cite{wang2018tienet}. 
The fundamental problem of such interpretability description is that these models implicitly assume neural network's understanding of abstraction levels similar to that of a radiologist.
Unlike radiologists, deep networks have no understanding of abstraction levels.
We assume, similar to computer vision, the network is getting higher activations at distinguishing textures without radiological context~\cite{nguyen2015deep} in heat maps and learning association rules without context in joint embedding.

From the interpretation point of view, upper levels of abstraction significantly reduce the uncertainty compared to the lower levels.
Let us take an example where a radiologist is explaining her diagnosis about breast cancer to another radiologist.
She can point out to a particular irregular calcification at a certain location represented by a set of pixels in the radiograph with the implicit assumption that her colleague uses similar abstractions to understand the anatomy - even if the colleague is trained at a different continent.
Her decision at the pixel level is interpretable (even if it does not agree to her colleague) because of one-to-one mappings at upper abstraction levels with little uncertainty.
Both of them agreed upon in a hierarchical fashion that the radiograph contains the image of a breast as well as its location, orientation, delineation and sub-parts.
These higher-level interpretations significantly reduce their uncertainty about the final decision, while focusing at the set of calcification pixels.        
Such a communication would be impossible with an alien radiologist who does not understand human anatomy in similar levels of abstraction - the uncertainty at the pixel level is simply too massive.
This exact scenario happens with a deep network.
Taking the heat map example, the deep network would fail to generate heat maps at all abstraction levels (e.g. anatomy, its subparts) unless explicitly trained to do so.

We argue that restricting ourselves into diagnostic interpretation has one major advantage over the general purpose interpretation - namely the diagnostic process has well defined levels of abstraction. 
This means a diagnostic model can have a complete mathematical definition removed from subjective biases. 
We theorize \emph{Diagnostic Interpretation} as the communication between two models, where \emph{one model is querying and adjusting its own decision process at all levels of abstraction to maximally emulate the decision process of the other model.}
One immediate benefit of this theory is that the usual question of accuracy-interpretability tradeoff does not arise here.  
In fact, within this theory, any “bad model” with low prediction accuracy can interpret another complex and superior model -- as long as the models share levels of abstractions.
A real-world example of this is the process when a trainee radiologist queries an expert and adjusts his model by emulating the expert's.        



%
%
%

Though the community has been broadly curious about defining the terms such as \emph{interpretability}, \emph{explainability} in semi-formal ways, formally defining the process of \emph{interpretation} and \emph{diagnostic interpretation} is often overlooked.
In section~\ref{Theory}, we formalize \emph{complete interpretation}, \emph{interpretability}, \emph{diagnostic interpretation}, \emph{confidence on interpretation} and \emph{$\epsilon$-interpretation}.
We derive an algorithm in section~\ref{Algo} and show an experimental evaluation in section~\ref{results}. 
Section~\ref{Discussion} discusses the possible impacts of the theory introduced here and concludes with some future direction.

\section{Theory Formulation}\label{Theory}
Let us assume two models $A$ and $B$ with overall representations $F_A$ and $F_B$. 
These models take similar images $S$ as input and predicts diagnosis label ($Y$)  as output i.e. $F_A:S \rightarrow Y_A$ and $F_B:S \rightarrow Y_B$.
Note that this description is devoid of ground truth, the prediction accuracy of each model for unseen examples is irrelevant for this theory.
  
We propose a \textbf{strong assumption} about interpretation process. 
$A$ and $B$ can only communicate for interpretation if both of them contains exactly same levels of abstractions i.e. if $\{{}^k F_A\}_{k=1,2..,K}$ and $\{{}^k F_B\}_{k=1,2..,K}$ are representation of $A$ and $B$ at $K$ different abstraction levels, then these levels must have a one-to-one relationship.
At each level $k$, $A$ predicts as follows: ${}^k F_A: S \rightarrow {}^k Y_A$. 
$B$ follows the same suite.

This assumption is derived based on the observation discussed in section~\ref{intro}.
This observation also suggests significant lowering of uncertainty at upper levels of abstraction, though the exact relationship is unknown.
Since most of the complex modern learning models (e.g. CNN, random forest) fall flat in this category compared to the diagnostic models of radiologists, we propose directions to relax this assumption in the definition of $\epsilon$-interpretation.

\begin{definition}{Complete interpretation: }\label{CI}
	A complete interpretation is the process of communication through exhaustive querying by the known model $A$ to minimize it's uncertainty about the decision boundary of the target black-box model $B$. 
	
	The exhaustive querying ensures no further information gain even if querying continues indefinitely longer.
\end{definition}


\begin{definition}{Interpretability: }\label{interp}
	Interpretability ($I_{A \leftarrow B}$) is the ratio between information gain about target model $B$'s decision boundary through~\emph{complete interpretation} and the initial uncertainty about $B$'s decision boundary. 
	
	More formally,
	\begin{equation} \label{interpretability}
	I_{A \leftarrow B} = \dfrac{H_{A \not\leftarrow B} - H_{A \leftarrow B}}{H_{A \not\leftarrow B}}
	\end{equation}
	
\end{definition}


Here, $H_{A \not\leftarrow B}$ represents the entropy about $B$'s decision boundary (across all $k$), before the process of interpretation starts. 
$H_{A \leftarrow B}$, on the other hand, represents the entropy (across all $k$) after complete interpretation. 
Assuming each abstraction level is independent of the others, overall entropy $H_{A \leftarrow B}$ across all levels of abstraction can be calculated as the sum of individual entropies at each abstraction level i.e.

\begin{equation}\label{entropy}
	H_{A \leftarrow B} = \sum_{k=1}^K {}^{k}H_{A\leftarrow B} ({}^{k}Y_A,{}^{k}Y_B)
\end{equation}

where ${}^{k}H_{A\leftarrow B}$ is the entropy at abstraction level $k$.
Based on how $I_{A \leftarrow B}$ approaches the two extremes, there are two boundary cases: either (a) model $B$ would be interpreted completely when $H_{A \leftarrow B} = 0$ which in turn means $I_{A \leftarrow B} = 1$ or (b) it would be impossible to interpret if $H_{A \leftarrow B} =  H_{A \not\leftarrow B}$  which in turn means $I_{A \leftarrow B} = 0$.

\begin{figure*}[t!]
	\centering
	\centerline{\includegraphics[width=0.7\textwidth]{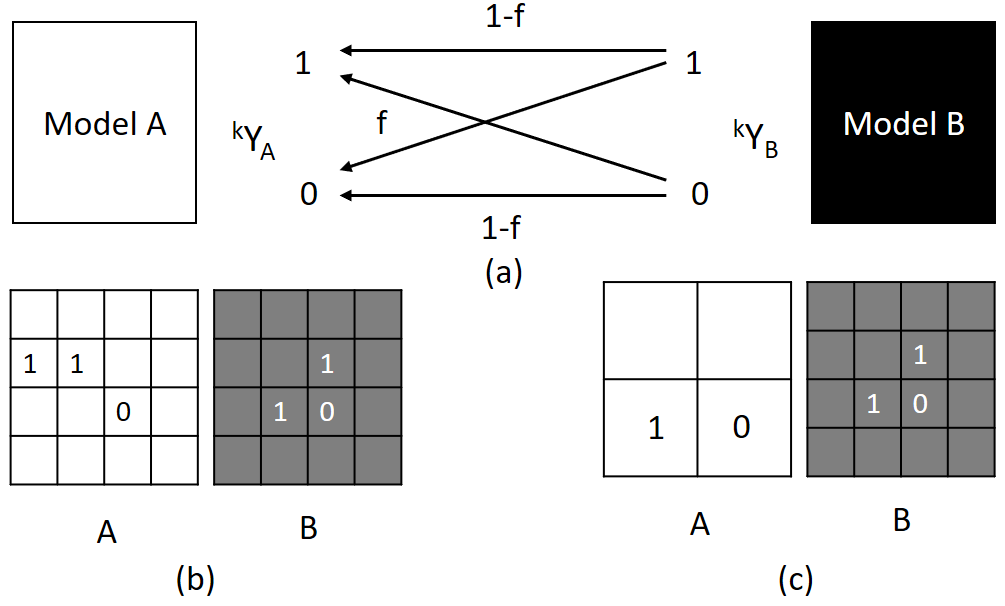}}
	\caption{(a) Interpretation of a binary classification problem can be thought of as communication through symmetric binary channels. Two sets of rules for square binary images of size $4 \times 4$ with same level of abstraction yet of similar (b) or different complexity(c). Grids for models $A$ and $B$ (b,c) show necessary entries for pixel values to predict $Y=1$. Empty pixels mean the model ignores those for prediction.}
	\label{fig:Example_Full_Interp}
\end{figure*}

\textbf{Toy Example: }Let's consider a simple example of square binary images of size $4 \times 4$. 
The possible set $\hbar$ of all such images has the massive cardinality $ ||\hbar|| = 2^{16}$.
Both model $A$ and $B$ are represented at a single scale $K=1$, with decision mechanism of similar complexity as shown in figure~\ref{fig:Example_Full_Interp}(b).   

For such an example, it is possible to do a \emph{Complete interpretation}. 
Let's compute the two terms in RHS of equation~\ref{interpretability} separately for this example.
If we initialize $F_{A \not\leftarrow B}$ with $F_A$, then the interpretation process looks like a symmetric binary channel, where model $A$ and $B$ does not conclude at the same prediction with probability $f$, as shown in figure~\ref{fig:Example_Full_Interp}(a).
This means the presence of an initial entropy $h$ that can be calculated as following: $h = f log_2\dfrac{1}{f} + (1-f) log_2\dfrac{1}{1-f}$.

Since, an exhaustive search in this case ensures lossless representation of $F_B$ in $F_{A \not\leftarrow B}$,  $H_{A \leftarrow B} = 0$, which means $I_{A \leftarrow B} = 1$.
Now, let's consider a less complex decision mechanism $F_A$ that can only approximate the decision process of $B$ (figure~\ref{fig:Example_Full_Interp}(c)). 
In such scenario, if initial entropy is $h$ and final entropy is $H_{A \leftarrow B} = h'$, the interpretability $I_{A \leftarrow B} = \dfrac{h-h'}{h}$ i.e. $0<I_{A \leftarrow B}<1$.
This means, the decision boundary of $B$ can be interpreted with $\dfrac{h-h'}{h}$ certainty w.r.t. $A$. 
Note that, though same level of abstraction ensures complete interpretation, the complexity of the model means $I_{A \leftarrow B}$ might or might not be 1.

\textbf{Practical Limitations of Complete interpretation: }The problem of definition~\ref{CI} is the potentially massive cardinality of $\hbar$.
If $F_A$ and $F_B$ are defined in the image space, the massive uncertainty on average about $B$'s decision render complete interpretation impractical since the process requires querying through and reasoning about all the possible images, resulting in an expensive optimization problem.

We relax the definition~\ref{CI} into a practical process of \emph{diagnostic interpretation} by exploiting two intuitions.
The \textbf{first intuition} is that even though $||\hbar||$ is massive, the image manifold where realistic images lie and radiological decisions are made is much lower dimensional.  
The \textbf{second intuition} comes from social science observations.
These observations suggest that humans only need a few example cases for reasonable interpretation~\cite{miller2017explanation}.

Equipped with these two intuitions, we relax definition~\ref{CI} into realistic definition of \emph{diagnostic interpretation} by minimal querying on image manifold for maximal information gain. 


\begin{definition}{Diagnostic interpretation: }\label{DI}
Diagnostic interpretation is the process of querying minimal number of times $T$ by $A$ to maximize the interpretability ($I^T_{A \leftarrow B}$) of $B$. 

More formally,

\begin{equation} \label{D_interp}
I^T_{A \leftarrow B} = \min_{S \subseteq \hbar} -\sum_{s^t \in S} I^t_{A \leftarrow B} + \lambda||S||
\end{equation}
	
\end{definition}

Parameter $\lambda$ balances between interpretability and number of queries whereas $||S||$ represents the cardinality of the subset of images from which query images are sampled.
$I^t_{A \leftarrow B}$ for each $t$ can be calculated as:

\begin{equation} \label{I_t}
	I^t_{A \leftarrow B} = \dfrac{H_{A \not\leftarrow B} - H^t_{A \leftarrow B}(Y^t_A,Y^t_B|s^t)}{H_{A \not\leftarrow B}}
\end{equation}

The entropy for each sample across all the levels of abstraction can be calculated as:

\begin{equation} \label{H_tk}
	H^t_{A \leftarrow B}(Y^t_A,Y^t_B|s^t) = \sum_{k=1}^K {}^{k}H^t_{A\leftarrow B} ({}^{k}Y_A^t,{}^{k}Y_B^t|s^t)
\end{equation}

Based on the values of entropy, three extremal cases are possible: (a) after $t$-queries, model $B$ would be interpreted completely when $H^t_{A \leftarrow B} = 0$ which in turn would lead to $I^t_{A \leftarrow B} = 1$, (b) between queries, entropy would not change even after model update i.e. $H^{t-1}_{A \leftarrow B} =  H^t_{A \leftarrow B}$ which means $I^t_{A \leftarrow B} = 0$ and finally (c) initial entropy is 0 i.e. $H^0_{A \not\leftarrow B}=0$ which means model $B$ and $A$ are exactly same from an information perspective, and already interpretability is 1.

\begin{figure*}[t!]
	\centering
	\centerline{\includegraphics[width=0.7\textwidth]{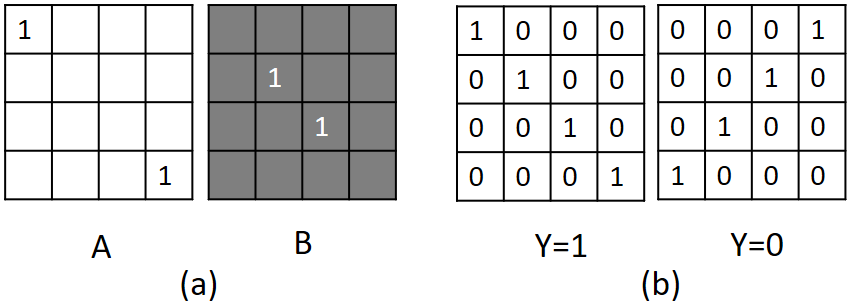}}
	\caption{Second toy example of two different rules (a) for classifying diagonals (b).}
	\label{fig:Example_Diag_Interp}
\end{figure*}

\textbf{Toy Example: }Let's consider the simple example of square binary images of size $4 \times 4$ introduced earlier.
However, the underlying image manifold this time is the set of diagonal images and the decision problem is to classify the two diagonals, as shown in figure~\ref{fig:Example_Diag_Interp}(b).
Two models $A$ and $B$ for this classification decision is shown in figure~\ref{fig:Example_Diag_Interp}(a).
In this example, though only two 'real images' are possible, the image space has $2^{16}$ possibilities.
We can marginalize this massive image space in multiple ways by exploiting the~\emph{first intuition}.
We have considered an intuitive relaxed envelop of single pixel flips to represent noise.
There are 32 (=16 $\times$ 2) such cases, resulting in a total of 34 (= 2+32) possible images.

Of these 34 images, there are only 4 images where $A$ disagrees with $B$ i.e. $P(Y_A \neq Y_B)_{A \not\leftarrow B} = \dfrac{4}{34}$.
Plugging in entropy equation, we get $H_{A \not\leftarrow B} = h = -0.52$.
We consider all those 4 cases and update the rules of A, to get final $F_{A \leftarrow B} = F_B$ and $H_{A \leftarrow B} =0$ accordingly.
So, interpretability $I_{A \leftarrow B} = 1$.

This view of diagnostic interpretation suggests that accuracy-interpretability tradeoff is not real i.e. it is unnecessary to simplify a model for interpretation by trading accuracy.
In fact, a \emph{model of any complexity can be diagnostically interpretable} as long as it respects the levels of abstraction of the source model.


\textbf{Limitations of Diagnostic interpretation: }In a typical CAD scenario where known model $A$ (e.g. decision process of radiologist) tries to interpret black-box model $B$ (e.g. DNN or kernelized learning), the strong assumption regarding one-to-one mapping of abstraction levels behind the definition~\ref{DI} does not hold anymore.  
Without levels of abstraction assumption, the necessary way to ensure Diagnostic interpretation is the exhaustive search of $\hbar$ - the absence of which should reduce confidence about the quality of interpretation.
As the first step toward relaxation of the strong assumption, we introduce \emph{Confidence on Interpretation} ($\epsilon$) in the following.

\begin{definition}{Confidence on Interpretation: }\label{Degree}
The Confidence on Interpretation ($\epsilon$) is the ratio of cardinality of the subset from which images are sampled for interpretation and the cardinality of the set of all possible images that the models might encounter. 	
\end{definition}

Confidence on Interpretation measures typicality of a sample on which the interpretation is performed i.e. how likely the models might encounter such a sample.
Without any explicit assumption about the image manifold, lower bound of Confidence on Interpretation $\epsilon| = \dfrac{||S||}{||\hbar||}$.
Note that, in real world situation where the image space is massive, neglecting the \emph{Confidence on Interpretation} might lead to severe consequences, for example wrong calculation of interpretabiliy, abundance of unexplainable examples etc.


Equipped with \emph{Confidence on Interpretation}, we have a relaxed and practical definition of interpretability in deep learning context. 

\begin{definition}{$\epsilon$-interpretation: }\label{EPS}
$\epsilon$-interpretation is the process of querying minimal number of times $T$ by the model A to maximize the $\epsilon$-interpretability (${}^{\epsilon}I_{A \leftarrow B}$).	
	More formally,
	
	\begin{equation} \label{EPS_Interp}
	{}^{\epsilon}I^T_{A \leftarrow B} = \min_{S \subseteq \hbar} -\sum_{s^t \in S} {}^{\epsilon}I^t_{A \leftarrow B} + \lambda||S||
	\end{equation}
	
\end{definition}

That is in the absence of correspondence in levels of abstractions, after $t$ iterations of the $\epsilon$-interpretation process, we are at most $\epsilon$ confident about the interpretability ${}^{\epsilon}I^T_{A \leftarrow B}$ of model $B$ by model $A$.
$\epsilon$-interpretability per sample (${}^{\epsilon}I^t_{A \leftarrow B}$) can be calculated by the formulation derived in equation~\ref{H_tk} with $K=1$.


%
%


%
%
%
%
%
%
%
%

\section{Diagnostic Interpretation Algorithm}\label{Algo}
%
This algorithm can calculate either diagnostic interpretability, or under minor modification, $\epsilon$-interpretability.

\begin{algorithm}[H]
	\label{alg:DI}
	\SetAlgoLined
	\KwResult{$I^T_{A \leftarrow B}$ }
	$F^0_{A \leftarrow B} = F_A$\; 
	$H^0_{A \leftarrow B} = H_{A \not\leftarrow B}$\; 
	\While{$t < T$}{
		Sample $s^t \in S$ s.t. $Y^{t-1}_{A \leftarrow B} (s^t) \not= Y_{B} (s^t)$\;
		\For{each abstract level of abstraction $k$}{
			${}^{k}F^{t}_{A \leftarrow B} = {}^{k}F^{t-1}_{A \leftarrow B} + \delta {}^{k}F_{A}$ \;
			${}^{k}Y^{t}_{A \leftarrow B} = {}^{k}F^{t}_{A \leftarrow B} (s^t)  $ \;
		}
		Update $H^t_{A \leftarrow B}$ using equation~\ref{H_tk} \;
	}
	Calculate ${}^{\epsilon} I^T_{A \leftarrow B}$ using equation~\ref{EPS_Interp} \;
	\caption{Diagnostic Interpretation algorithm}
\end{algorithm}

Here, $\delta {}^{k}F_{A}$ is the update rule of model $A$ at each level of abstraction.
Note that both sampling and update rule is not specified in algorithm~\ref{alg:DI} - these are free parameters that can be chosen based on problem assumptions and $A$.


\section{Evaluation}\label{results}
%
%
%

We evaluate algorithm~\ref{alg:DI} for calculating ${}^{\epsilon}I^t_{A \leftarrow B}$ of a deep neural network ($B$) with respect to a linear SVM ($A$).

\begin{figure*}[t!]
	\centering
	\centerline{\includegraphics[width=\textwidth]{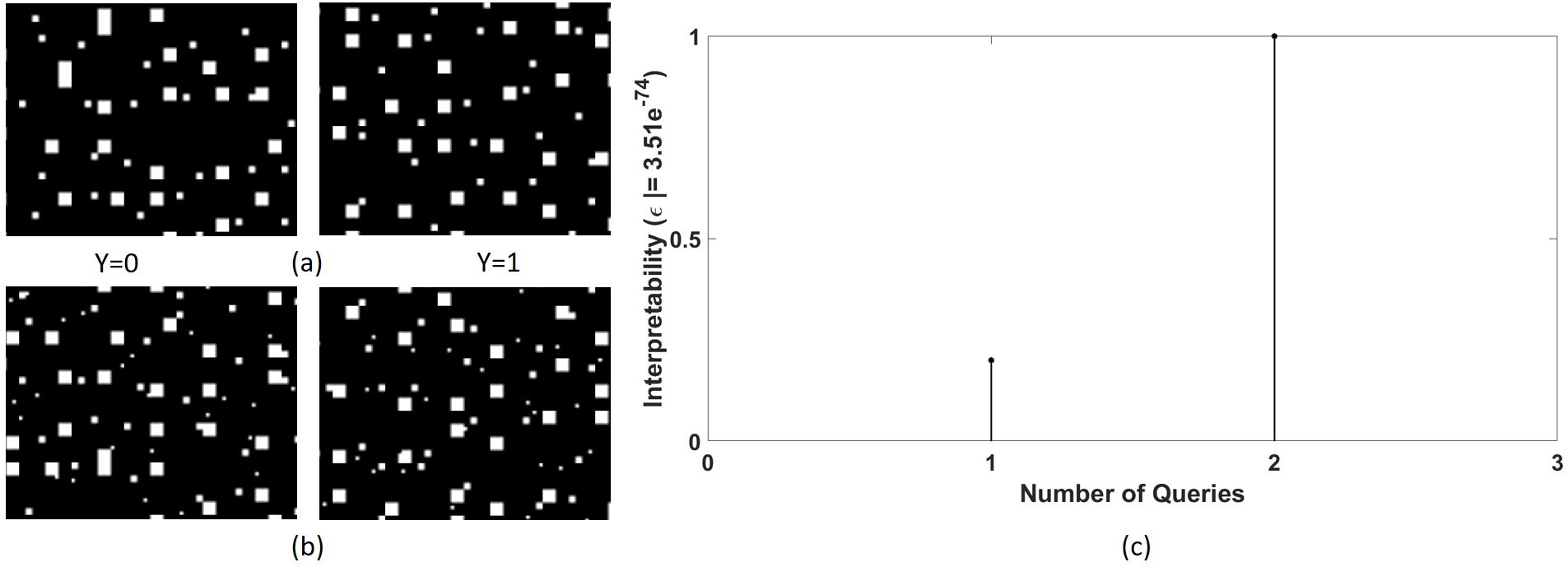}}
	\caption{Given a binary classification problem for binary images (a,b), how ${}^{\epsilon}I^t_{A \leftarrow B}$ evolved over number of queries (c).}
	\label{fig:T_vs_I}
\end{figure*}

\textbf{Dataset: }We consider a simple problem - binary classification of binary images.
The images with left square smaller than the right are assigned label $Y=0$ and the opposite images are assigned $Y=1$ (see Figure~\ref{fig:T_vs_I}(a)).
We also consider all the images with 1 pixel flipped as envelop, belonging to the corresponding class (Figure~\ref{fig:T_vs_I}(a)) 

\textbf{Interpretation: }
We trained a simple convolutional neural network (model $B$) of the following architecture: C32-C32-F32 with $3 \times 3$ convolutional kernels having ReLu activations followed by MaxPooling layers, using binary crossentropy loss.
We trained a simple liner SVM as model $A$ to interpret model $B$.
200 randomly sampled images (100 from each group) are used for training, while $\epsilon$-interpretation was performed on the dataset of $4068(=2^9 + 2^{16})$ images i.e. all normal images and 1-pixel flip envelop.
Based on algorithm~\ref{alg:DI}, we consider random sampling of images where model $A$ and $B$ disagrees.
As a simple update rule, we created an intermediate training dataset by concatenating the sampled image along with its label as predicted by CNN - linear SVM model was re-trained on this intermediate dataset.

For this simple dataset, we trained CNNs 10 different times and accumulated the average result in figure~\ref{fig:T_vs_I}(c). 
By the second iteration, all 10 test runs achieved ${}^{\epsilon}I^t_{A \leftarrow B} = 1$.
However, note how flimsy the confidence of interpretation lower bound ($\epsilon| = \dfrac{4068}{2^{256}} = 3.51e^{-74}$) is for this problem, even after considering the 1-pixel flip envelop.

\section{Discussion}\label{Discussion}
The main goal of this paper is to start a discourse on the possibility of quantitative analysis and understanding of diagnostic interpretability.
Looking at interpretation under the prism of communication between two models, we propose some basic definitions and an algorithm to emulate the process of interpretation for calculating interpretability.
Neither the definitions nor the algorithm is complete - in fact the algorithm might need to be improved and the definitions need to be revisited in future. 
This theory predicts that the classical idea of interpretability-accuracy tradeoff is true in a limited sense
In fact the real bottleneck might be the maximum achievable unique decodability of $B$ w.r.t. $A$'s levels of abstractions.
We can safely assume that in future, complex yet highly interpretable models can be designed, as long as such models respect $A$'s levels of abstractions.

From a human computer interaction perspective, it would be an interesting future direction to study whether the theoretically calculated interpretability has any correlation to radiologist's perception.
It would also be interesting to study how many images, on average, a radiologist look at before trusting the black-box model.
Finally, for some form of deep learning to be acceptable as a clinical CAD system (either sime- or fully-automatic), gaining trust of the radiologist is essential.
Looking away from subjective bias might be a good first step toward gaining that trust.

\section{Acknowledgement}
Special thanks to David K{\"u}gler for vigorous discussions and cleaning up of initial formulation.
A thanks also goes to Johannes Fauser for an initial discussion.
Thanks to Arjan Kuijper and Salome Kazeminia for proof reading, commenting and checking the formulations.

\renewcommand{\bibsection}{\section*{References}} 
\bibliographystyle{splncs03}
\begingroup
\bibliography{Diag_Interp}
\endgroup

\end{document}